\documentclass[runningheads]{llncs}
\usepackage[title]{appendix}
\usepackage{blindtext}
\usepackage{hyperref}
\usepackage{breakurl}
\usepackage{graphicx}
\usepackage{epstopdf}
\usepackage{algpseudocode}
\usepackage{algorithm}
\usepackage{wrapfig}
\usepackage{enumitem}
\usepackage{csquotes}
\usepackage{nicefrac}
\usepackage{dcolumn}
\usepackage{float}
\usepackage{microtype}  
\usepackage{booktabs}
\usepackage{amsmath,amssymb} % define this before the line numbering.
\usepackage{color}
\usepackage{subfig}
\usepackage{multirow}
\usepackage{chngpage}
\newcommand*{\figref}[2][]{%
    \hyperref[{fig:#2}]{%
        Figure~\ref*{fig:#2}%
        \ifx\\#1\\%
        \else
        \,#1%
        \fi
    }%
}
\newcommand*{\methodref}[2][]{%
    \hyperref[{method:#2}]{%
        Method~\ref*{method:#2}%
        \ifx\\#1\\%
        \else
        \,#1%
        \fi
    }%
}
\begin{document}
\title{Towards Privacy-Preserving Visual Recognition via Adversarial Training: A Pilot Study}
% Replace with your title

\titlerunning{Privacy-Preserving Visual Recognition via Adversarial Training}
% Replace with a meaningful short version of your title
%
\author{Zhenyu Wu\inst{1} \and
Zhangyang Wang\inst{1} \and
Zhaowen Wang\inst{2} \and 
Hailin Jin\inst{2}}
%
%Please write out author names in full in the paper, i.e. full given and family names. 
%If any authors have names that can be parsed into FirstName LastName in multiple ways, please include the correct parsing, in a comment to the volume editors:
%\index{Lastnames, Firstnames}
%(Do not uncomment it, because you may introduce extra index items if you do that, we will use scripts for introducing index entries...)
\authorrunning{Zhenyu Wu, Zhangyang Wang, Zhaowen Wang and Hailin Jin}
% Replace with shorter version of the author list. If there are more authors than fits a line, please use A. Author et al.
%

\institute{Texas A\&M University, College Station TX 77843, USA
\email{\{wuzhenyu\_sjtu,atlaswang\}@tamu.edu} \\ \and
Adobe Research, San Jose CA 95110, USA \\ \email{\{zhawang,hljin\}@adobe.com}}
\maketitle              % typeset the header of the contribution

\begin{abstract}
This paper aims to improve privacy-preserving visual recognition, an increasingly demanded feature in smart camera applications, by formulating a unique adversarial training framework. The proposed framework explicitly learns a degradation transform for the original video inputs to optimize the trade-off between target task performance and the associated privacy budgets on the degraded video. A significant challenge is that the privacy budget, often defined and measured in task-driven contexts, cannot be reliably indicated using any single model performance because strong privacy protection has to sustain against any possible model that tries to hack privacy information. Such an uncommon situation has motivated us to propose two strategies, i.e., budget model \textbf{restarting} and \textbf{ensemble}, to enhance the generalization of the learned degradation on protecting privacy against unseen hacker models. Novel training strategies, evaluation protocols, and result visualization methods have been designed accordingly. Two experiments on privacy-preserving action recognition, with privacy budgets defined in various ways, manifest the compelling effect of the proposed framework in simultaneously maintaining high target task (action recognition) performance while suppressing the privacy breach risk. The code is available at \url{https://github.com/VITA-Group/Privacy-AdversarialLearning}.
\keywords{Visual privacy, adversarial training, action recognition}
\end{abstract}
\section{Introduction}
Smart surveillance or smart home cameras, such as Amazon Echo and Nest Cam, are now found in millions of locations to remotely link users to their homes or offices, providing monitoring services to enhance security and/or notify environment changes, as well as lifelogging and intelligent services. Such a prevalence of smart cameras has reinvigorated the privacy debate since most of them require to upload device-captured visual data to the centralized cloud for analytics. 

This paper explores how to make sure that those smart computer vision devices are only seeing the things that we want them to see (and how to define what we want)? Is it at all possible to alleviate privacy concerns without compromising user convenience?

At first glance, the question itself is posed as a dilemma: we would like a camera system to recognize important events and assist daily human life by understanding its videos while preventing it from obtaining sensitive visual information (such as faces) that can intrude people's privacy. Classical cryptographic solutions secure communication against unauthorized access from attackers. However, they are not immediately applicable to preventing authorized agents (such as the back-end analytics) from the unauthorized abuse of information, causing privacy breach concerns. The popular concept of differential privacy has been introduced to prevent an adversary from gaining additional knowledge by inclusion/exclusion of a subject, but not from gaining knowledge from released data itself \cite{cormode2010individual}. In other words, an adversary can still accurately infer sensitive attributes from any sanitized sample available, which does not violate any of the (proven) properties of differential privacy \cite{hamm2017minimax}. It thus becomes a new and appealing problem to find an appropriate transform on the collected raw visual data at the local camera end so that the transformed data itself will only enable specific target tasks while obstructing other undesired privacy-related tasks.
Recently, some new video acquisition approaches~\cite{butler2015privacy,dai2015towards,ryoo2017privacy} proposed to intentionally capture or process videos in extremely low-resolution to create privacy-preserving ``anonymized videos'', and showed promising empirical results.

In contrast, we formulate the privacy-preserving visual recognition in a unique adversarial training framework. The framework explicitly optimizes the trade-off between target task performance and associated privacy budgets by learning active degradations to transform the video inputs. We investigate a novel way to model privacy budget in a task-driven context. Unlike the standard adversarial training where two individual models compete, our framework's privacy budget cannot be simply defined with one single model. The ideal protection of privacy has to be universal and model-agnostic, i.e., obstructing every possible model from predicting privacy information. To resolve the so-called \textbf{``$\forall$ challenge''}, we propose two strategies, i.e., restarting and ensembling budget model(s), to enhance the generalization capability of the learned degradation to defend against unseen models. Novel training strategies and evaluation protocols have been proposed accordingly. Two experiments on privacy-preserving action recognition, with privacy budgets defined in different ways, manifest the proposed framework's effectiveness. With many problems left open and a considerable improvement room existing, we hope this pilot study will attract more community interests.
\section{Related Work}
\subsection{Privacy Protection in Computer Vision}

With pervasive cameras for surveillance or smart home devices, privacy-preserving visual recognition has drawn increasing interests from both industry and academia, mainly for two reasons. First, due to the devices' computationally demanding nature, it is often impractical to run visual recognition tasks at the resource-limited local device end. Communicating (part of) data to the cloud is indispensable. Second, while traditional privacy concerns mostly arise from the unsecured channel between cloud and device (e.g., malicious third-party eavesdropping), customers now possess increasing concerns against sharing their private visual information to the cloud (which might turn malicious itself). 

A few cryptographic solutions \cite{erkin2009privacy,yonetani2017privacy} were developed to encrypt visual information in a homomorphic way locally, i.e., the cryptosystems allow for basic arithmetic classifiers over encrypted data. However, many encryption-based solutions will incur high computational costs at local platforms. It is also challenging to generalize the cryptosystems to more complicated classifiers. \cite{chattopadhyay2007privacycam} combined the detection of regions of interest and the real encryption techniques to improve privacy while allowing general surveillance to continue. A seemingly reasonable and computationally cheaper option is to extract and transmit feature descriptors from raw images and transmit those features only. Unfortunately, a previous study \cite{mahendran2016visualizing} revealed that considerable information of original images could still be recovered from standard HOG or SIFT features (even they look visually distinct from natural images), making them fragile to privacy hacking too. 

An alternative toward a privacy-preserving vision system concerns the concept of anonymized videos. Such videos are intentionally captured or processed to be in special low-quality conditions, which only allow for recognizing some target events or activities, while avoiding the unwanted leak of the identity information for the human subjects in the video \cite{butler2015privacy,dai2015towards,ryoo2017privacy}. Typical examples of anonymized videos are videos made to have extremely low resolution (e.g., $16 \times 12$) by using low-resolution camera hardware \cite{dai2015towards}, based on image operations like blurring and superpixel clustering \cite{butler2015privacy}, or introducing cartoon-like effects with a customized version of mean shift filtering \cite{winkler2014trusteye}. \cite{pittaluga2015privacy,pittaluga2017pre} proposed to use privacy-preserving optics to filter sensitive information from the incident light-field before sensor measurements are made, by $k$-anonymity and defocus blur. Earlier work \cite{jia2014using} explored privacy-preserving tracking and coarse pose estimation using a network of ceiling-mounted time-of-flight low-resolution sensors. \cite{tao2012privacy} adopted a network of ceiling-mounted binary passive infrared sensors. However, both works handled only a limited set of activities performed at specific constrained areas in the room.  Later, \cite{ryoo2017privacy} showed that even at the extremely low resolutions, reliable action recognition could be achieved by learning appropriate downsampling transforms, with neither unrealistic activity-location assumptions nor extra specific hardware resources. The authors empirically verified that conventional face recognition easily failed on the generated low-resolution videos. The usage of low-resolution anonymized videos \cite{dai2015towards,ryoo2017privacy} is computationally cheaper and compatible with sensor and bandwidth constraints. However, \cite{dai2015towards,ryoo2017privacy} remain empirical in protecting privacy. In particular, neither were their models learned towards protecting any visual privacy nor were the privacy-preserving effects carefully analyzed and evaluated. In other words, privacy protection in \cite{dai2015towards,ryoo2017privacy} came as a ``side product'' of down-sampling, and was not a result of any optimization. The authors of \cite{dai2015towards,ryoo2017privacy} also did not extend their efforts to studying deep learning-based recognition, making their task performance less competitive. 

Very recently, a few learning-based approaches have come into play to ensure better privacy protection. \cite{sokolic2017learning} defined a utility metric and a privacy metric for a task entity, and then designed a data sanitization
function to achieve privacy while providing utility. However, they considered only simple sanitization functions such as linear projection and maximum mean discrepancy transformation. 

In \cite{raval2017protecting}, the authors proposed a game-theoretic framework between an obfuscator and an attacker, in order to hide visual secrets in the camera feed without significantly affecting the functionality of the target application. This seems to be the most relevant work to the proposed one: however, \cite{raval2017protecting} only discussed a toy task to hide QR codes while preserving the overall structure of the image. Another relevant work \cite{hamm2017minimax} addressed the optimal utility-privacy tradeoff by formulating it as a min-diff-max optimization problem. Nonetheless, The empirical quantification of privacy budgets in existing works \cite{sokolic2017learning,raval2017protecting,hamm2017minimax} only considered to protect privacy against \textit{one hacker model}, and was thus insufficient, for which we will explain more in Section 3.1.

\subsection{Privacy Protection in Social Media and Photo Sharing}
User privacy protection is also a topic of extensive interests in the social media field, especially for photo sharing. The most common means to protect user privacy in an uploaded photo is to add empirical obfuscations, such as blurring, mosaicing, or cropping out certain regions (usually faces) \cite{li2017blur}. However, extensive research showed that such an empirical means could be easily hacked too \cite{oh2016faceless,mcpherson2016defeating}. 

The latest work \cite{oh2017adversarial} described a game-theoretical system in which the photo owner and the recognition model strive for antagonistic goals of dis-/enabling recognition, and better obfuscation ways could be learned from their competition. However, it was only designed to confuse one specific recognition model, via finding its ``adversarial perturbations'' \cite{nguyen2015deep}. That can cause obvious overfitting as merely changing to another recognition model will likely put the learning efforts in vain: such perturbations even cannot protect privacy from \textit{human eyes}. Their problem setting thus deviated far away from our target problem. Another notable difference is that we usually hope to cause minimum perceptual quality loss to those photos after applying any privacy-preserving transform to them in social photo sharing. The same concern does not exist in our scenario, allowing us to explore much more free, even aggressive image distortions.

A useful resource to us was found in \cite{orekondy17iccv}, which defined concrete privacy attributes and correlated them to image content. The authors categorized possible private information in images, and then run a user study to understand privacy preferences. They then provided a sizable set of 22k images annotated with 68 privacy attributes, on which they trained privacy attribute predictors.

\subsection{Recognition from Visually Degraded Data}

One crucial challenge to enable the usage of anonymized videos is to ensure reliable performance of the target tasks on those lower-quality videos, besides suppressing the undesired privacy leak. Among all low visual quality scenarios, visual recognition in low resolution is probably best studied. \cite{CVPR16,liu2017enhance,cheng2017robust} showed that low-resolution object recognition could be significantly enhanced through proper pre-training and domain adaption. Low-resolution action recognition has also drawn growing interests: \cite{ryoo2017extreme} proposed a two-stream multi-Siamese CNN that learns the embedding space to be shared by low-resolution videos down-sampled in different ways, on top of which a transform-robust action classifier was trained. \cite{chen2016semi} leveraged a semi-coupled filter-sharing two-stream network to learn a mapping between the low- and high-resolution feature space. 

In comparison, the ``low-quality'' anonymized videos in our case are generated by learned and more complicated degradations, other than simple downsampling \cite{CVPR16,chen2016semi}. 
\section{Technical Approach}
\subsection{Problem Definition}
Assume our training data $X$ (raw visual data captured by camera) are associated with a target task $\mathcal{T}$ and a privacy budget $\mathcal{B}$. We mathematically express the goal of privacy-preserving visual recognition as below ($\gamma$ is a weight coefficient):
\begin{equation}
\begin{array}{l}\label{privacy}
\min_{f_T, f_d} L_T(f_T(f_d(X)), Y_T) + \gamma L_B (f_d(X)),
\end{array}
\end{equation} 
where $f_T$ denotes the model to perform the target task $\mathcal{T}$ on its input data. Since $\mathcal{T}$ is usually a supervised task, e.g., action recognition or visual tracking, a label set $Y_T$ is provided on $X$, and a standard cost function $L_T$ (e.g., softmax) is defined to evaluate the task performance on $\mathcal{T}$. On the other hand, we need to define a budget cost function $L_B$ to evaluate the privacy leak risk of its input data: the larger $L_B$, the higher privacy leak risk. Our goal is to seek such an \textit{active degradation} function $f_d$ to transform the original $X$ as the common input for both $L_T$ and $L_B$, such that:
\begin{itemize}
\item The target task performance $L_T$ is minimally affected compare to when using the raw data, i.e., $\min_{f_T, f_d} L_T(f_T(f_d(X)), Y_T) \approx \min_{f'_T} L_T(f'_T(X), Y_T)$.
\item The privacy budget $L_B$ is greatly suppressed compared to raw data, i.e., $L_B(f_d(X)) \ll L_B(X)$.
\end{itemize}
The definition of the privacy budget cost $L_B$ is not straightforward. Practically, it needs to be placed in concrete application contexts, often in a task-driven way. For example, in smart workplaces or smart homes with video surveillance, one might want to avoid disclosing persons' faces or identities. Therefore, to reduce $L_B$ could be interpreted as to suppress the success rate of identity recognition or verification on the transformed video $f_d(X)$. Other privacy-related attributes, such as race, gender, or age, can be similarly defined too. 
We denote the privacy-related annotations (such as identity label) as $Y_B$, and rewrite $L_B (f_d(X))$ as $L_B (f_b(f_d(X)), Y_B)$, where $f_b$ denotes the budget model to predict the corresponding privacy information. {\em{Different from $L_T$}}, minimizing $L_B$ will encourage $f_b(f_d(X))$ to diverge from $Y_B$ as much as possible.

Such a \textit{supervised, task-driven} definition of $L_B$ poses at least two-fold challenges: (1) the privacy budget-related annotations, denoted as $Y_B$, often have less availability than target task labels. Specifically, it is often challenging to have both $Y_T$ and $Y_B$ ready on the same $X$; (2) considering the nature of privacy protection, it is not sufficient to merely suppress the success rate of one $f_b$ model. Instead, define a privacy prediction function family $\mathcal{P}$: $f_d(X) \rightarrow Y_B$, the ideal privacy protection of $f_d$ should be reflected as \textbf{suppressing every possible model} $f_b$ from $\mathcal{P}$. \textit{That diverts from the common supervised training goal}, where one only needs to find one model to successfully fulfill the target task. We re-write the general form (\ref{privacy}) with the task-driven definition of $L_B$:
\begin{equation}
\begin{array}{l}\label{task_privacy}
\min_{f_T, f_d} L_T(f_T(f_d(X), Y_T) + \gamma \max_{f_b \in \mathcal{P}} L_B (f_b(f_d(X)), Y_B).
\end{array}
\end{equation} 
%It constitutes an intriguing ``$\exists$-$\forall$'' optimization problem: 
For the solved $f_d$, the two goals should be simultaneously satisfied: (1) there \textbf{exists} (``$\exists$'') at least one $f_T$ function that can predict $Y_T$ from $f_d(X)$ well; (2) \textbf{for all} (``$\forall$'') $f_b$ functions $\in \mathcal{P}$, \underline{none of them} (even the best one) can reliably predict $Y_B$ from $f_d(X)$. Most existing works chose an empirical $f_d$ (e.g., simple downsampling) and solved $\min_{f_T} L_T(f_T(f_d(X), Y_T)$ \cite{dai2015towards,CVPR16}. \cite{ryoo2017privacy} essentially solved $\min_{f_T,f_d} L_T(f_T(f_d(X), Y_T)$ to jointly adapted $f_d$ and $f_T$, after which the authors empirically verified the effect of $f_d$ on $L_B$ (defined as face recognition error rates). Those approaches lack the explicit optimization towards privacy budgets, and thus have no guaranteed privacy-protection effects. 

\vspace{-0.5em}
\paragraph{Comparison to Standard Adversarial Training} The most notable difference between (\ref{task_privacy}) and existing works based on standard adversarial training \cite{raval2017protecting,oh2017adversarial} lies in whether the adversarial perturbations are optimized for ``fooling'' \textit{one specific} $f_b$, or \textit{all possible} $f_b$s. We believe the latter to be necessary, as it considers generalization ability to suppressing unseen privacy breach. Moreover, most existing works seek perturbations with minimal human visual impacts, e.g., by enforcing $\ell_p$ norm constraint on the pixel domain. That is unaligned with our purpose. Our model could be viewed as to minimize the perturbation in the (learned) feature domain of the target utility task.

\subsection{Basic Framework}
\begin{wrapfigure}{r}{0.5\textwidth}
\centering {
\includegraphics[width=0.5\textwidth]{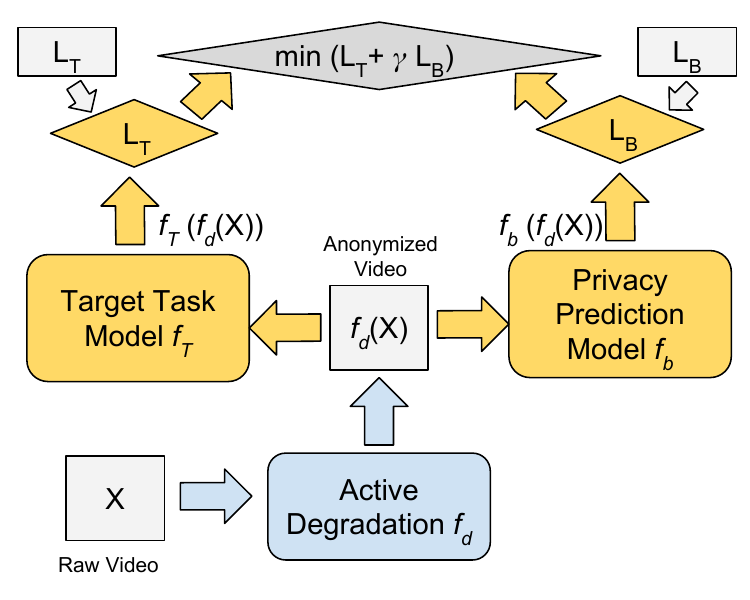}
}
\vspace{-2em}
\caption{Basic adversarial training framework for privacy-preserving visual recognition.} 
\vspace{-2.5em}
\label{fig:privacySR}
\end{wrapfigure}

\noindent \textbf{Overview} \figref{privacySR} depicts a model architecture to implement the proposed formulation (\ref{task_privacy}). It first takes the original video data $X$ as the input, and passes it through the active degradation module $f_d$ to generate the anonymized video $f_d(X)$. During training, the anonymized video simultaneously goes through a target task model $f_T$ and a privacy prediction model $f_b$. All three modules, $f_d$, $f_T$, and $f_b$, are learnable and can be implemented by neural networks. The entire model is trained under the hybrid loss of $L_T$ and $L_B$. By tuning the entire pipeline from end to end, $f_d(X)$ will find the optimal task-specific transformation to the advantage of the target task and the disadvantage of the privacy breach, fulfilling the goal of privacy-preserving visual recognition. After training, we can apply the learned active degradation at the local device (e.g., camera) to convert incoming video to its anonymized version, which is then transmitted to the backend (e.g., cloud) for target task analysis.

The proposed framework leads to an adaptive and end-to-end manageable pipeline for privacy-preserving visual recognition. Its methodology is related to the emerging research of feature disentanglement \cite{xiang2017linear}. That technique leads to non-overlapped groups of factorized latent representations, each of which would properly describe information corresponding to particular attributes of interest.

Previously it was applied to generative models \cite{desjardins2012disentangling,siddharth2016learning} and reinforcement learning \cite{higgins2017darla}.

Similar to GANs \cite{goodfellow2014generative} and other adversarial models, our training is prone to collapse and/or bad local minimums. We thus propose a carefully-designed training algorithm with a three-module alternating update strategy, explained \textbf{in the supplementary}, which could be interpreted as a three-party game. In principle, we strive to prevent any of the three modules ($f_d$, $f_T$, and $f_b$) to change ``too quickly''. Thus, keep monitoring $L_T$ and $L_b$ to decide which of the three modules to be updated next. 

\noindent \textbf{Choices of $f_d$, $f_T$, and $f_b$} The choices of the three modules will significantly impact the performance. As \cite{ryoo2017privacy} pointed out, $f_d$ can be constructed as a nonlinear mapping by filtering. The form of $f_d$ can be flexible, and its output $f_d(X)$ is unnecessary to be a natural image. For simplicity, we choose $f_d$ to be a ``learnable filtering'' in the form of a 2-D convolutional neural network (CNN), whose output $f_d(X)$ will be a 2-D feature map of the same resolution as the input video frame. Such a choice only facilitates the initial concatenation of building blocks, e.g., $f_T$ and $f_b$ often start with pre-trained models on natural images. 

Besides, $f_d(X)$ should preferably be in a compact form and light to transmit, considering it will be sent to the cloud through (limited-bandwidth) channels.

To ensure the effect of $f_d$, sufficiently strong $f_T$ and $f_b$ models should be chosen to compete with each other. We employ state-of-the-art video recognition CNNs for corresponding tasks, and adapt them for the degraded input $f_d(X)$ using the robust pre-training strategy proposed in \cite{CVPR16}. 

Particular attention should be paid towards the budget cost (second term) defined in (\ref{task_privacy}), which we refer to as \textbf{``the $\forall$ Challenge''}: if we use $f_b$ with some pre-defined CNN architecture, how could we be sure that it is the ``best possible'' privacy prediction model? In other words, given a $f_d$ function that manages to fail one $f_b$ model, is it possible that some other $f'_b \in \mathcal{P}$ would still be able to predict $Y_B$ from $f_d(X)$, thus leaking privacy? 
While it is computationally intractable to search over $\mathcal{P}$ exhaustively, a naive empirical solution would be to chose a robust privacy prediction model, hoping that a $f_d$ function can confuse this strong one will be able to fool other possible functions as well. However, the resulting $f_d(X)$ may still overfit the artifacts of one specific $f_b$ and fails to generalize. Section 3.3 will introduce two more advanced and feasible recipes. 
%$f_b$ = $\arg\max_{f_b \in \mathcal{P}} L_B (f_b(f_d(X)), Y_B)$

\noindent \textbf{Choices of $L_T$ and $L_B$} Without loss of generality, we assume both target task $f_T$ and privacy prediction $f_b$ to be classification models and output class labels. To optimize the target task $\mathcal{T}$'s performance, $L_T$ could be simply chosen as the KL divergence: $KL(f_T(f_d(X), Y_T)$. 

Choosing $L_B$ is non-standard and tricky since we require minimizing the privacy budget $L_B (f_b(f_d(X)), Y_B)$ to enlarge the divergence between $f_b(f_d(X))$ and $Y_B$. One possible choice is the negative KL divergence between the predicted class vector and the ground truth label, but minimizing a concave function will cause a ton of numerical instabilities (often explosions). Instead, we use the predicted class vector's negative entropy function and minimize it to encourage ``uncertain'' predictions. Meanwhile, we will use $Y_B$ to ensure a sufficiently strong $f_b$ at the initialization (see 4.1.2). Furthermore, $Y_B$ will play a critical role in model restarting (see 3.3).
\vspace{-1.0em}
\subsection{Addressing the $\forall$ Challenge}

To improve the generalization of learned $f_d$ over all possible $f_b \in \mathcal{P}$ (i.e., any model cannot reliably predict privacy), we hereby discuss two easy-to-implement and straightforward options. 

Other more sophisticated model re-sampling or model-search approaches, e.g., \cite{zoph2017learning}, will be explored in future work.

\textbf{Budget Model Restarting} At certain point of training (e.g., when the privacy budget $L_B (f_b(f_d(X)))$ stops decreasing any further), we replace the current weights in $f_b$ with random weights. Such random restarting aims to avoid trivial overfitting between $f_b$ and $f_d$ (i.e., $f_d$ is only specialized at confusing the current $f_b$) without incurring more parameters. We then start to train the new model $f_b$ to be a strong competitor, w.r.t. the current $f_d(X)$: specifically, we freeze the training of $f_d$ and $f_T$, and change to minimizing $KL(f_b(f_d(X)), Y_B)$, until the new $f_b$ has been trained from scratch to become a strong privacy prediction model over current $f_d(X)$. We then resume adversarial training by unfreezing $f_d$ and $f_T$, as well as replacing the loss for $f_b$ back to the negative entropy. It can repeat several times.

\textbf{Budget Model Ensemble} The other strategy proposes to approximate the continuous $\mathcal{P}$ with \textit{a discrete set of $M$ sample functions}. Assuming the budget model ensemble $\{f_b^i\}_{i=1}^M$, we turn to minimizing the following discretized surrogate of (\ref{task_privacy}):
\begin{equation}
\begin{array}{l}\label{ensemble}
\min_{f_T, f_d} L_T(f_T(f_d(X), Y_T) + \gamma \max_{i \in \{1, 2,..., M\}} L_B (f_b^i(f_d(X))). \\
\end{array}
\end{equation} 
At each iteration (mini-batch), minimizing (\ref{ensemble}) will only suppress the model $f_b^i$ with the largest $L_B$ cost, e.g., the ``most confident'' one about its current privacy prediction. The previous basic framework is a special case of (\ref{ensemble}) with $M$ = 1. The ensemble strategy can easily be combined with re-starting.

\subsection{Two-Fold Evaluation Protocol}
Apart from training data $X$, assume we have an evaluation set $X^e$, accompanied with both target task labels $Y_T^e$ and privacy annotations $Y_B^e$. Our evaluation is significantly more complicated than classical visual recognition problems. After applying the learned active degradation, we need to examine in two folds: (1) whether the learned target task model maintains satisfactory performance; (2) whether the performance of an \underline{arbitrary} privacy prediction model will deteriorate. The first one can follow the standard routine: applying the learned $f_d$ and $f_T$ to $X^e$, and computing the classification accuracy $A_T$ via comparing $f_T(f_d(X^e))$ w.r.t. $Y_T^e$: the higher the better.

The second evaluation is insufficient if we only observe that the learned $f_d$ and $f_b$ lead to poor classification accuracy on $X^e$, because of the $\forall$ challenge. In other words, $f_d$ needs to generalize not only in the data space, but also w.r.t. the $f_b$ model space. To empirically verify that $f_b$ prohibits reliable privacy prediction for other possible models, we propose a novel procedure: we first re-sample a different set of $N$ models $\{f_b^j\}_{j=1}^N$ from $\mathcal{P}$; none of them will have overlap with the $M$ budget models used in training. We then train each of them to predict privacy information, over the degraded \underline{training data $X$} by applying the learned $f_d$, i.e., minimizing $f_b^j(f_d(X))$, $j = 1, ..., N$. Eventually, we apply each trained $f_b^j$ and $f_d$ on $X^e$ and compute the $j$-th model classification accuracy. The highest accuracy achieved among the $N$ models on $f_d(X^e)$, denoted as $A_b^N$, will be by default used to indicate the privacy protection capability of $f_d$: the lower, the better. 
\vspace{-1.0em}
\section{Experiments}
\vspace{-0.5em}
We present two experiments on \textit{privacy-preserving action recognition}, as proofs-of-concept for our proposed general framework. We choose video-based action recognition for the target task because it is a highly demanded feature in many smart homes and smart workplaces. The definition of privacy will vary by contexts, and we will study two settings: (1) avoiding the leak of person identities present in the current work; and (2) avoiding the leak of multiple privacy attributes, e.g., multiple crowdsourced attributes studied in \cite{orekondy17iccv}. We emphasize that the generality of the proposed framework (\ref{task_privacy}) can fit in a wide variety of target task and privacy information compositions.

\subsection{Identity-Preserving Action Recognition on SBU}
\paragraph{Problem Setting} The SBU Kinect Interaction Dataset \cite{kiwon_hau3d12} is a two-person interaction dataset for video-based action recognition,  with 8 types of actions and 13 different actor pairs annotated.

We define action recognition as the target task $\mathcal{T}$, and the privacy budget task $\mathcal{B}$ as reducing the correct identification rates of the actor pairs in the same video. We note that the target trade-off is highly challenging to achieve. As can be seen from the first Table \textbf{in the supplementary}, the actor pair recognition task easily achieves over 98\% accuracy on the original dataset, and stands robust even when the frames are downsampled 28 times, while the action recognition performance already starts to deteriorate significantly. We compare the following five methods:
\begin{itemize}[wide, labelwidth=!, labelindent=0pt, noitemsep,topsep=0pt]
\item \textbf{Method 1} (naive downsampling): using raw RGB frames under different down-sampling rates \label{method:1}.
\item \textbf{Method 2} (proposed w/o re-starting): applying the proposed adversarial training to RGB frames, using budget model ensemble \textit{without} restarting \label{method:2}.
\item \textbf{Method 3} (proposed): applying  the proposed adversarial training to RGB frames, using budget model ensemble  \textit{with} restarting \label{method:3}.
\item \textbf{Method 4}: detecting and cropping out faces from RGB frames.
\item \textbf{Method 5}: detecting and cropping out whole actor bodies from RGB frames.
\end{itemize}
Method 1 follows \cite{ryoo2017privacy}, while Methods 4 and 5 are inspired by \cite{li2017blur}. 

\vspace{-0.5em}
\paragraph{Implementation Details} We segment video sequences into groups of 16 frames, and use those frame groups as our default input data $X$. 
 We use the C3D net \cite{tran2015learning} as the default action recognition model, i.e. $f_T$. For the $f_b$ identity recognition model, we choose MobileNet \cite{howard2017mobilenets} to identify actor pair in each frame, and use average pooling to aggregate the frame-wise predictions. The active degradation module $f_d$ adopts the image transformation network in \cite{Johnson2016Perceptual}.

We choose $\gamma = 2.0$ to suppress the identity recognition performance on SBU.  We first initialize the active degradation module $f_d$ as a reconstruction of the input. We next take the pre-trained version of C3D net and concatenate it with $f_d$, and jointly train them for action recognition on the SBU dataset, to initialize $f_T$. We then freeze them both, and start initializing $f_b$ (MobileNet) for the actor pair identification task, by adapting it to the output of the currently trained $f_d$. Experiments show that such initializations provide robust starting points for the follow-up adversarial training.
If budget model restarting is adopted, we set to ``restart'' MobileNet from random initialization after every 100 iterations. The number of ensemble budget models $M$ varies in $\{1,2,4,6,8,10,12,14,16,18\}$. Different budget models can be obtained via setting different depth-multiplier parameter \cite{howard2017mobilenets} of MobileNet.

\paragraph{Evaluation Procedure} We will follow the procedure described in Section 3.4, for two-fold evaluations on the SBU testing set. For the set of models used towards the privacy-protection examination, we sample $N$ = 10 popular image classification CNNs, a list of which can be found \textbf{in the supplementary}. Among them, 8 models start from ImageNet-pretrained versions, including MobileNet~\cite{howard2017mobilenets} (different from those used in training), ResNet~\cite{He2015} and Inception~\cite{szegedy2016rethinking}. To eliminate the possibility that the initialization might prohibit privacy prediction, we also intentionally try another 2 models trained from scratch (random initialization). We did not choose any non-CNN image classification model for two reasons: (1) CNNs have state-of-the-art performance and also strong fitting capability when re-trained; (2) most non-CNN image classification models rely on effective feature descriptors that are designed for natural images. Since $f_d(X)$/$f_d(X_e)$ are no longer natural images, the effectiveness of such models is in jeopardy too. 

\paragraph{Results and Analysis}
\begin{wrapfigure}{r}{0.5\textwidth}
	\centering{
	\includegraphics[width=0.5\textwidth]{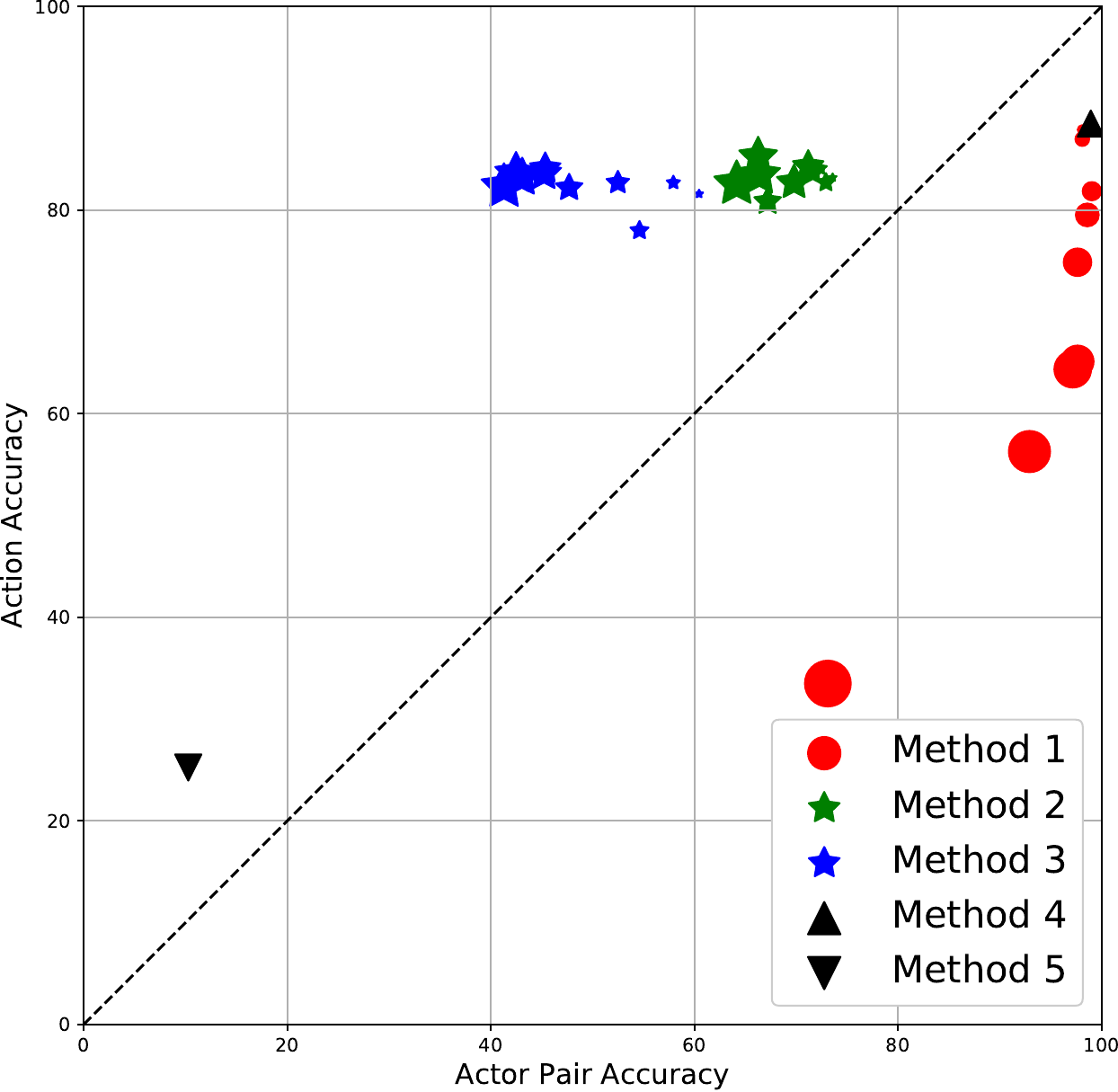}
    }
    \vspace{-2.0em}
	\caption{Target and Budget Task Performance Trade-off on SBU Dataset.} 
    \vspace{-2.5em}
	\label{fig:sbu_tradeoff}
\end{wrapfigure}
We present a \textbf{novel visualization} in \figref{sbu_tradeoff},to display the trade-off between the action recognition accuracy $A_T$ and the actor pair recognition accuracy $A_b^N$, in an easy-to-interpret way. All accuracy numbers for both task evaluation can be found \textbf{in the supplementary}. To read the figure, note that a desirable trade-off should incur a minimal loss of $A_T$ (y-axis) when reducing $A_b^N$ (x-axis). Therefore, a point \textbf{closer to the upper left corner} denotes a more desirable $f_d$ model that achieves better trade-off, since it incurs less utility performance loss (larger $A_T$) while suppressing more even the best of $N$ unseen privacy prediction models (smaller $A_b^N$). 

For Method 1, a larger marker (circle) size represents a larger downsampling rate. For Methods 2 and 3, a larger marker (star) size denotes more budget models used in the ensemble (i.e., larger $M$). Both Methods 4 and 5 give single points.
Observations can be summarized below:
\begin{itemize}[leftmargin=*]
\item Methods 2 and 3 have obvious advantages over naive downsampling (Method 1) in terms of a more compelling trade-off between $A_T$ and $A_b^N$. 
\item Method 3 achieves much better individual model performance than Methods 2, showing the significance of \textbf{re-starting} in overcoming over-fitting.
\item For Method 3 (similarly for Method 2), adding more $f_b$ models into the budget model ensemble (i.e., increasing $M$) consistently leads to points closer to the desired upper-left (despite small randomness caused by training). The results show that proposed \textbf{budget model ensemble} technique effectively alleviates the $\forall$ challenge, making $f_d$ generalize better to unseen $f_b$s. 
\item Neither Methods 4 nor 5 performs competitively. Cropping out faces does not add to the protection of identity privacy (its $A_b^N$ is almost the same as using raw RGB frames without downsampling), since body shape and clothing styles reveal significant amounts of identity information too. Cropping out the body removes not only identity information but also deteriorates action recognition.  
\end{itemize}

In brief, Method 3 (the proposed, with both restarting and ensemble) is able to achieve most compelling results on SBU, the best ($M$ = 18) at $A_T$ = 82.3\% and $A_b^N$  = 41.8\%, without relying on extra hardware (e.g., depth camera). A few visualized examples of learned $f_d(X)$ can be found in \figref{sbu_visualization}.

\subsection{Protecting Multiple Privacy Attributes on UCF-101/VISPR}
\paragraph{Problem Setting} The UCF101 dataset \cite{soomro2012ucf101} is a more challenging action recognition benchmark with 101 different action classes and much richer scenes than SBU. The newly-released Visual Privacy (VISPR) dataset \cite{orekondy17iccv} is a collection of 22, 167 images annotated with 68 privacy attributes, e.g., semi-nudity, hobbies, face, race, gender, skin color, and so on. 

Either alone cannot be used for our adversarial training because neither dataset has a pair of competing tasks/labels. However, we show the possibility of jointly utilizing UCF101 and VISPR to train one $f_d$ function, such that it simultaneously enables effective action recognition (as measured on UCF-101) and suppresses multiple privacy attribute prediction (as measured on VISPR), via a novel cross-dataset training and evaluation strategy. 

\vspace{-0.5em}
\paragraph{Cross-Dataset Training and Evaluation} $f_d$ and $f_T$ are the same as the SBU one. Regarding $f_b$ for the task of multiple privacy attribute prediction, we adopt budget models from \{ResNet-V1-50, ResNet-V2-50, MobileNet-1.0, MobileNet-0.75\} and vary the number of $M$ in $\{1,2,3,4\}$. They are trained with a multi-label classification loss with sigmoid activations, identically to \cite{orekondy17iccv}. $\gamma$ is chosen as 0.5. Rather than recognition accuracy, the multiple privacy attribute prediction is evaluated by Class-based Mean Average Precision (C-MAP), whose definition can be referred to \cite{orekondy17iccv}.

The major hassle is the unavailability of $\mathcal{T}$ and $\mathcal{B}$ on the same dataset: since UCF-101 does not have privacy attributes annotated, we cannot directly perform adversarial training and evaluate privacy protection on it; similarly for VISPR. We notice that \cite{orekondy17iccv} trained the model on VISPR to be a privacy predictor for general images. We also visually observe that the VISPR model can correctly detect privacy attributes in UCF-101 videos (examples \textbf{in the supplementary}). Therefore, we hypothesize that the privacy attributes have the right ``transferability'' between UCF-101 and VISPR. Thus, we use a privacy prediction model trained on VISPR to assess the privacy leak risk on UCF-101.

Instead of using all 68 attributes in \cite{orekondy17iccv}, we find that many of them rarely appear in UCF-101 (shown \textbf{in the supplementary}). We thus create two subsets for training and evaluating budget models here: one \textit{VISPR-17} set consists of 17 attributes that occur most in UCF-101 and their associated images in VISPR; the other \textit{VISPR-7} set is further a subset of VISPR-17, that include 7 privacy attributes out of 17 that are most common in smart home settings. Their attribute lists are \textbf{in the supplementary}. 

During training, we have two pipelines: one is $f_d$ + $f_T$ trained on UCF-101 for action recognition; the other is $f_d$ + $f_b$ trained on VISPR to suppress multiple privacy attribute prediction. The two pipelines \textit{share the same parameters} for $f_d$. The initialization and alternating training strategy remain unchanged from SBU. During the evaluation, we perform the first part of the two-fold evaluation, e.g., action recognition, on the UCF-101 testing set. We then evaluate the performance of the $N$-model examination on privacy protection, using the VISPR-17/7 testing sets. Such cross-dataset training and evaluation shed on new possibilities on training privacy-preserving recognition models, even under the practical shortages of datasets that have been annotated for both tasks.

\begin{figure}
    	\vspace{-1em}
	\centering {
		\includegraphics[width=350pt]{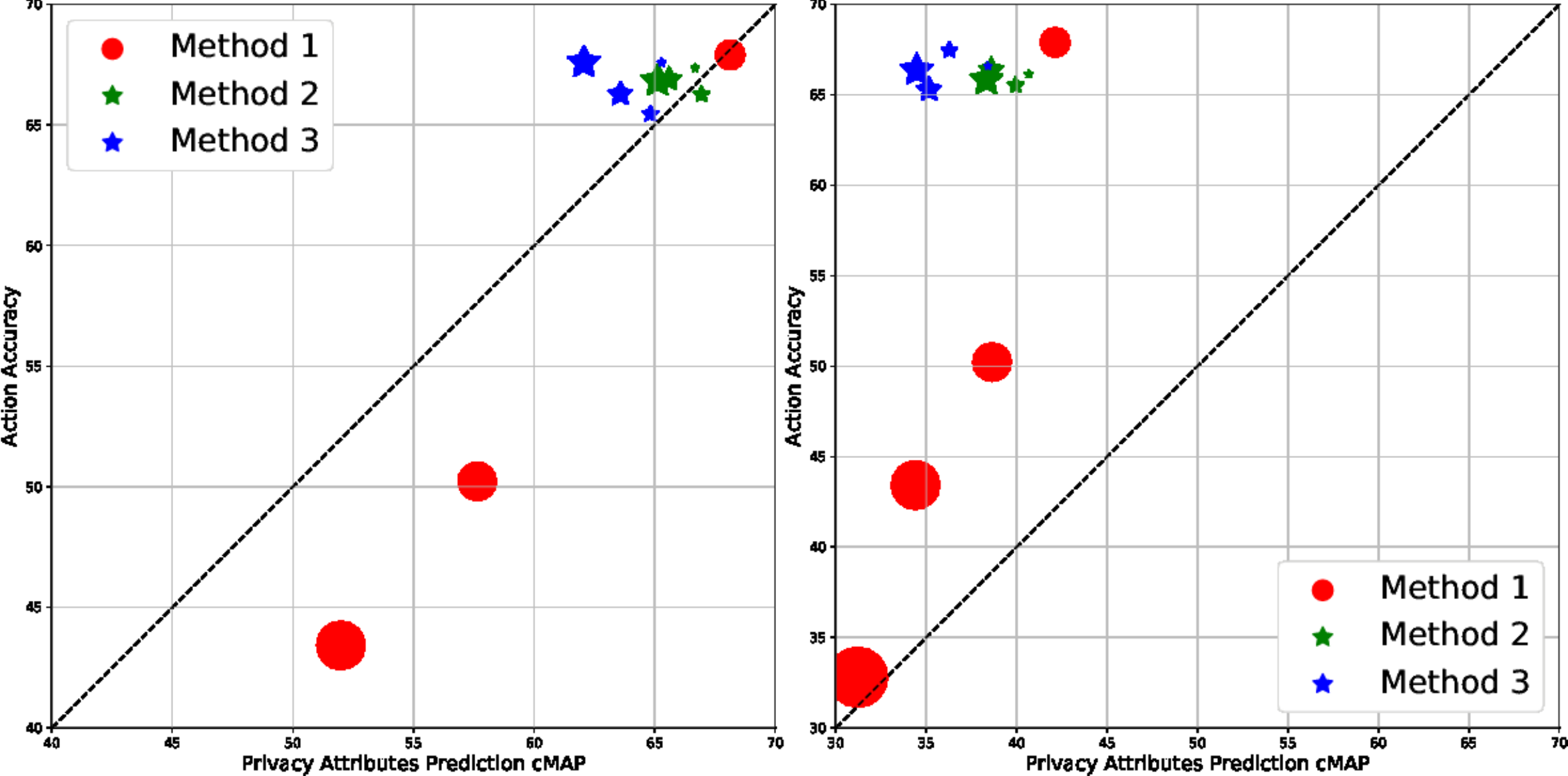}
	}
	    \vspace{-2.0em}
	\caption{Performance Trade-off on UCF-101/VISPR dataset. The left one is on VISPR-17 and the right one on VISPR-7.} 
        	\vspace{-1em}
	\label{fig:vispr_tradeoff}
        \vspace{-0.5em}
\end{figure}

        %	\vspace{-0.5em}
\paragraph{Results and Analysis} We choose Methods 1, 2, and 3 for comparison, defined the same as SBU. All the quantitative results, as well as visualized examples of $f_d(X)$ on UCF-101, are shown \textbf{in the supplementary}. Similarly to the SBU case, simply downsampling video frames (even with the aid of super-resolution as we tried) will not lead to any competitive trade-off between action recognition (at UCF-101) and privacy prediction suppression (at VISPR). 
As is shown in \figref{vispr_tradeoff}, our proposed adversarial training again leads to more favorable trade-offs on VISPR-17 and VISPR-7, with major conclusions concur with SBU: both ensemble and restarting help $f_d$ generalize better against privacy breach. 

\begin{figure}[!htb]
  \centering\includegraphics[width = 1.2in]{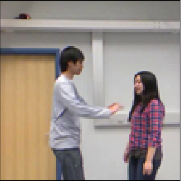}
  \caption*{Original RGB Frame from UCF-101 (Label: Pushing)}
  	\label{fig:sbu_original}
\captionsetup[subfigure]{labelformat=empty}
\begin{tabular}{cccc}
\subfloat[Method 2, M=1]{\includegraphics[width = 1.2in]{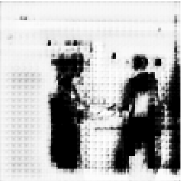}} &
\subfloat[Method 2, M=4]{\includegraphics[width = 1.2in]{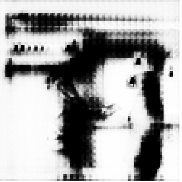}} &
\subfloat[Method 2, M=8]{\includegraphics[width = 1.2in]{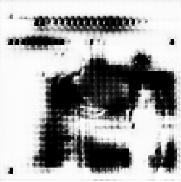}} &
\subfloat[Method 2, M=14]{\includegraphics[width = 1.2in]{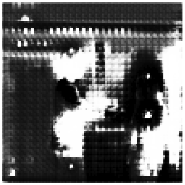}}\\
\subfloat[Method 3, M=1]{\includegraphics[width = 1.2in]{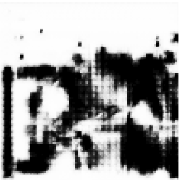}} &
\subfloat[Method 3, M=4]{\includegraphics[width = 1.2in]{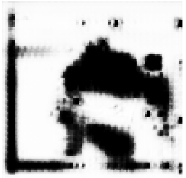}} &
\subfloat[Method 3, M=8]{\includegraphics[width = 1.2in]{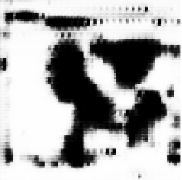}} &
\subfloat[Method 3, M=14]{\includegraphics[width = 1.2in]{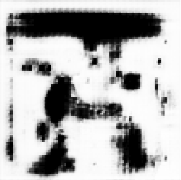}}
\end{tabular}
    \vspace{-0.5em}
\caption{Example frames after applying the learned degradation on SBU.}
	\label{fig:sbu_visualization}
    \vspace{-1.8em}
\end{figure}

\vspace{-1em}
\vspace{-0.5em}
\section{Limitations and Discussions}
\vspace{-0.5em}

One anonymous reviewer noted that a possible alternative to avoid leaking visual privacy to the cloud is to perform action recognition completely at the local device. In comparison, our proposed solution is motivated by at least three folds: \textbf{i)} for a single utility task (which is not just limited to action recognition), running $f_d$ on the device is much more compact and efficient than full $f_T$ For example, our $f_T$ model (11-layer C3D net) has over 70 million parameters. In contrast, $f_d$ is a much more compact 3-layer CNN with 1.3 million parameters. At the inference, the total time cost of running $f_T$ over the SBU testing set is 45 times more than running $f_d$. It also facilitates upgrading to more sophisticated $f_T$ models; \textbf{ii)} The smart home scenario calls for the scalability of multiple utility tasks (computer vision functions). It is not economical to load all utility models in the device. Instead, we can train one $f_d$ to work with multiple utility models and only store and run $f_d$ at the device.
%, which has better task scalability, 
More utility models (if no overlap with privacy) could be possibly added in the cloud by training on $f_d(X)$; \textbf{iii)} We further point out that the proposed approach can further have a broader practical application scope beyond the smart home, e.g., \underline{de-identified data sharing}. 

The current pilot study is preliminary in many ways, and there is a large performance room to improve until achieving practical usefulness. First, the definition of $\mathcal{B}$ and $L_B$ is core to the framework. Considering the $\forall$ challenge, the current budget model ensemble is a rough discretized approximation of $\mathcal{P}$. More elegant ways to tackle this $\forall$ optimization can lead to further breakthroughs in universal privacy protection. Second, adversarial training is well-known to be difficult and unstable. Improved training tricks, such as \cite{salimans2016improved}, will be useful. 

Third, a lack of related benchmark datasets, on which  $\mathcal{T}$ and $\mathcal{B}$ are both appropriately defined, has become a bottleneck. 
We see that more concrete and precise privacy definitions, such as VISPR attributes, can undoubtedly result in better feature disentanglement and $\mathcal{T}$-$\mathcal{B}$ performance trade-offs. Current cross-dataset training and evaluation partially alleviate the absence of dedicated datasets. However, the inevitable domain mismatch between two datasets can still hurdle the performance. We plan to refer to crowdsourcing to identify and annotate privacy-related attributes on existing action recognition or other benchmarks, which we hope could help promote this research direction.

\begin{subappendices}
\renewcommand{\thesection}{\Alph{section}}%
% or try \arabic{section}
\section{Adversarial Training Algorithm}

Algorithm~\ref{alg:AdvAlg} outlines a complete and unified adversarial training algorithm using the ensemble of $M$ budget models, with restarting. If we choose $M$ = 1 and skip the restarting step, it is reduced to the basic adversarial training framework. 

The algorithm could also be viewed as a 3-competitor game: $f_d$ as an \textbf{obfuscator}, $f_b$ (or the ensemble) as an \textbf{attacker}, and $f_T$ as an \textbf{utilizer}. Algorithm~\ref{alg:AdvAlg} then essentially solves the following two optimization problems iteratively (single $f_b$ case for example):
%The problem in our paper can also be formulated as a game among 3 competitors: obfuscator, attacker and utility module, which can be mathematically formulated as: 
\begin{align}
&\min_{f_d, f_T} L_T(f_T(f_d(X)), Y_T) - \gamma H (f_b(f_d(X))), \label{eq:1} \\
&\min_{f_b \in \mathcal{P}} L_B(f_b(f_d(X)), Y_B). \label{eq:2} 
\end{align}
% \begin{equation}
% \begin{array}{l}
% \vspace{-0.5em}
% \text{(Problem a)} \min_{f_d, f_T} L_T(f_T(f_d(X)), Y_T) - \gamma H (f_b(f_d(X))), \newline
% \end{array}
% \begin{array}{l}
% \vspace{-0.5em}
% \text{(Problem b)} \min_{f_b \in \mathcal{P}} L_B(f_b(f_d(X)), Y_B),
% \end{array}
% \end{equation}
where both $L_T$ and $L_B$ are softmax functions, $H$ is the entropy function. In the $M$-ensemble case, (\ref{eq:2}) will search for the worst case to minimize. 
\vspace{-1em}
\begin{algorithm}
\caption{Adversarial Training for Privacy-Preserving Visual Recognition.}
\label{alg:AdvAlg}
\begin{algorithmic}
\State{Given pre-trained active degradation module $f_d$, target task module $f_T$, and $M$ budget modules \{$f_b^1,\cdots f_b^M$\}} 
\For{number of training iterations}
\State{Sample a mini-batch of {k} examples \{$X_1, \cdots, X_k$\}}
\State{Update \textbf{active degradation module} $f_d$ (weights $w_d$) with stochastic gradients:}\\ \Comment{Suppress only the most confident one among all $M$ budget models} \\
 \Comment{The L1 Loss term is only used in the SBU experiment} \\
 \Comment{The $L_b$ is negative entropy loss}
\State{ $\nabla_{w_d} \frac{1}{k}\displaystyle\sum_{j=1}^k [L_T(f_T(f_d(X_j)), Y_{Tj})+\gamma\displaystyle\max_{i \in \{1,\cdots, M\}}{L_b(f_b^i(f_d(X_j)))} + \alpha ||f_d(X_j)-X_j||_1]$} \\
\While{target task validation accuracy $\leq$ Threshold$_1$} \\
\Comment{Threshold$_1$ = 90\% for SBU and 70\% for UCF101/VISPR}
\State{Sample mini-batch of $k$ examples \{$X_1, \cdots, X_k$\}}
\State{Update \textbf{target task module} $f_T$ (weights $w_T$) and \textbf{active degradation}} 
\State{\textbf{module} $f_d$ (weights $w_d$), with stochastic gradients:}
\\ \Comment{Avoid too weak competitor on the $f_T$ side.}
\State{$\nabla_{w_T} \frac{1}{k} \displaystyle\sum_{j=1}^kL_T(f_T(f_d(X_j)), Y_{Tj})$, $\nabla_{w_d} \frac{1}{k} \displaystyle\sum_{j=1}^kL_T(f_T(f_d(X_j)), Y_{Tj})$}
\EndWhile
\While{budget task training accuracy $\leq$ Threshold$_2$} \\ \Comment{Threshold$_2$ = 95\% for both datasets}
\State{Sample mini-batch of k examples \{$X_1, \cdots, X_k$\}}
\State{Update \textbf{budget task module} $f_b$ (weights $w_b$) by stochastic gradients:}
\\ \Comment{Avoid too weak competitor on the $f_b$ side.}
\State{$\nabla_{w_b}\frac{1}{k}\displaystyle\sum_{j=1}^k{\displaystyle\sum_{i=1}^M{L_b(f_b^i(f_d(X_j)),Y_{Bj})}}$}
 \Comment{The $L_b$ is cross-entropy loss}
\EndWhile
\If {current training iteration \% 100 = 0} \\ \Comment {We empirically restart all $M$ budget models every 100 iterations}
\State {Restart all $M$ budget models, and repeat Algorithm 1 from the beginning.}
%\State %\Goto \texttt{marker}
%Repeat Algorithm 1 from the beginning.
\EndIf
\EndFor
\end{algorithmic}
\end{algorithm}

\section{Experiments on SBU}
\subsection {Results for Methods 1}

The proposed identity-preserving action recognition task on SBU is a very challenging one, since videos are taken in highly controlled indoor environments and all actors are clearly viewable in the central regions of each frame. The identity recognition task can also utilize information other than faces: the body shape and even clothes colors are invariant for the same actor across different videos/actions. Different actors wear very distinct clothes with different colors and textures. Table \ref{table:low-res-sbu} displays the trade-off numbers at different downsampling ratios $s$, for Methods 1.

\subsection{Two-Fold Evaluation Results for Methods 2 and 3}

Table \ref{table:sbu-2fold-eval} displays the details numbers, for the second part of our proposed two-fold evaluation, with $N$ = 10 models. The top sub-table is for Method 2, and the bottom sub-table for Method 3. 

The corresponding action recognition results, i.e. the first part of two-fold  evaluation, are also attached after either sub-Table. 

We want to make an additional note here: for Methods 1, 4 and 5, the privacy prediction is evaluated using \underline{only one model}; while in Methods 2 and 3, the privacy suppression effect is evaluated using the \underline{highest achievable number} among $N$ = 10 different models. Therefore, the evaluation protocol for Methods 2 and 3 is \textbf{``stricter''}, and its gain on privacy protection compared to Methods 1, 4, 5 will be essentially \textbf{``underestimated''}, if we just directly compare accuracy numbers.

\subsection{Visualization Examples of Learned Degradations on SBU}
Please refer to \figref{sbu_visualization} for visualized examples of learned $f_d(X)$.

\begin{figure}[htbp]
  \centering\includegraphics[width = 1.2in]{figures/sbu_visualization/04.png}
  \caption*{Original RGB Frame from UCF-101 (Label: HandStandPushup)}
\captionsetup[subfigure]{labelformat=empty}
\begin{tabular}{cccc}
\subfloat[Method 2, M=1]{\includegraphics[width = 1.2in]{figures/sbu_visualization/NoResample_1_04.png}} &
\subfloat[Method 2, M=2]{\includegraphics[width = 1.2in]{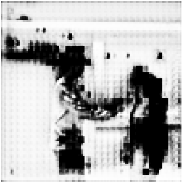}} &
\subfloat[Method 2, M=4]{\includegraphics[width = 1.2in]{figures/sbu_visualization/NoResample_4_04.png}} &
\subfloat[Method 2, M=6]{\includegraphics[width = 1.2in]{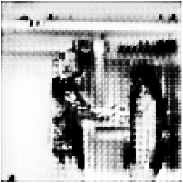}} \\
\subfloat[Method 3, M=1]{\includegraphics[width = 1.2in]{figures/sbu_visualization/Resample_1_04.png}} &
\subfloat[Method 3, M=2]{\includegraphics[width = 1.2in]{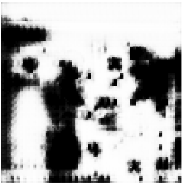}} &
\subfloat[Method 3, M=4]{\includegraphics[width = 1.2in]{figures/sbu_visualization/Resample_4_04.png}} &
\subfloat[Method 3, M=6]{\includegraphics[width = 1.2in]{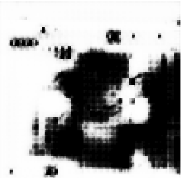}} \\ 
\subfloat[Method 2, M=8]{\includegraphics[width = 1.2in]{figures/sbu_visualization/NoResample_8_04.png}} &
\subfloat[Method 2, M=10]{\includegraphics[width = 1.2in]{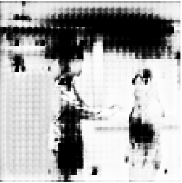}} &
\subfloat[Method 2, M=12]{\includegraphics[width = 1.2in]{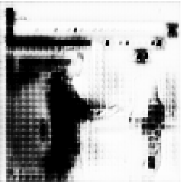}} &
\subfloat[Method 2, M=14]{\includegraphics[width = 1.2in]{figures/sbu_visualization/NoResample_14_04.png}} \\
\subfloat[Method 3, M=8]{\includegraphics[width = 1.2in]{figures/sbu_visualization/Resample_8_04.png}} &
\subfloat[Method 3, M=10]{\includegraphics[width = 1.2in]{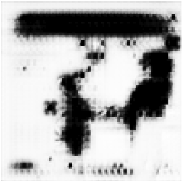}} &
\subfloat[Method 3, M=12]{\includegraphics[width = 1.2in]{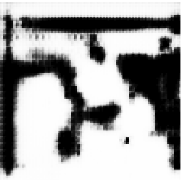}} &
\subfloat[Method 3, M=14]{\includegraphics[width = 1.2in]{figures/sbu_visualization/Resample_14_04.png}} \\ 
\end{tabular}
\caption{Example frames after applying the learned degradation on SBU}
	\label{fig:sbu_visualization}
\end{figure}

\begin{table}[htbp]
\begin{adjustwidth}{-.5in}{-.5in}
\begin{center}
\caption{The action recognition and actor pair recognition accuracies w.r.t. the spatial downsampling ratio $s$, using pre-trained C3D net and MobileNet.} % title of Table
\centering % used for centering table
\begin{tabular}{|c|c|c|c|c|c|c|c|c|c|c|c|} % centered columns (4 columns)
\hline %inserts double horizontal lines
{} & {} & s=1 & s=2 & s=3 & s=4 & s=6 & s=8 & s=14 & s=16 & s=28 & s=56\\
%heading
\hline
\multirow{2}{*}{Method 1} & Action & 88.83 & 87.90 & 86.98 & 81.86 & 79.53 & 74.88 & 65.12 & 64.37 & 56.28 & 33.49 \\ [1ex] % [1ex] adds vertical space
\cline{2-12}
(RGB Downsampling) & Actor & 98.87 & 97.23 & 96.45 & 95.50 & 95.24 & 94.11 & 93.94 & 92.15 & 90.28 & 60.93\\ [1ex]
\hline %inserts single line
\end{tabular}
\label{table:low-res-sbu} % is used to refer this table in the text
\end{center}
\end{adjustwidth}
\end{table}

\begin{table}[htbp]
\begin{adjustwidth}{-.5in}{-.5in}
\begin{center}
\caption{SBU Two Fold Evaluation} % title of Table
%\centering % used for centering table
\begin{tabular}{|c|c|c|c|c|c|c|c|c|c|c|} % centered columns (4 columns)
\hline %inserts double horizontal lines
{} & M=1 & M=2 & M=4 & M=6 & M=8 & M=10 & M=12 & M=14 & M=16 & M=18 \\ [0.5ex] % inserts table
%heading
\hline % inserts single horizontal line
resnet\_v1\_50 & 70.8 & 65.4 & 70.3 & 67.2 & 65.1 & 68.3 & 65.8 & 61.7 & 62.4 & 59.3 \\ % inserting body of the table
resnet\_v1\_101  & 68.3 & 67.6 & 71.4 & 69.4 & 66.8 & 69.7 & 63.0 & 62.5 & 59.2 & 57.0 \\
resnet\_v2\_50 & 62.6 & 62.1 & 61.9 & 64.9 & 63.3 & 62.3 & 58.4 & 61.1 & 62.9 & 60.8 \\
resnet\_v2\_101 & 69.6 & 66.9 & 71.4 & 68.9 & 66.1 & 64.2 & 65.2 & 64.9 & 64.8 & 60.0\\
mobilenet\_v1\_100 & \textbf{73.6} & 71.8 & \textbf{72.9} & 65.4 & 65.7 & \textbf{71.2} & 67.5 & 65.4 & \textbf{67.3} & 63.2 \\
mobilenet\_v1\_075  & 71.3 & \textbf{72.4} & 71.4 & 70.9 & 66.5 & 66.3 & 66.1 & \textbf{66.3} & 65.5 & 61.1 \\
inception\_v1 & 66.7 & 60.8 & 66.4 & 58.9 & 64.2 & 60.5 & 58.5 & 61.8 & 57.4 & 63.5 \\
inception\_v2& 60.6 & 61.3 & 68.7 & 67.6 & 60.3 & 59.1 & 62.3 & 61.1 & 61.6 & 62.1 \\ 
mobilenet\_v1\_050\textsuperscript{$\ddagger$} & 71.2 & 70.5 & 69.6 & \textbf{71.6}& \textbf{67.2} & 70.6 & 67.5 & 65.2 & 64.4 & 63.2 \\
mobilenet\_v1\_025\textsuperscript{$\ddagger$} & 70.6 & 71.5 & 71.9 & 70.2 & 66.4 & 70.7 & \textbf{69.8} & 65.8 & 65.5 & \textbf{64.2} \\ [1ex]
\hline %inserts single line
C3D  & 83.2 & 84.1 & 82.7 & 83.6 & 80.8 & 88.3 & 82.7 & 83.3 & 83.5 & 82.6 \\ [1ex] % [1ex] adds vertical 
\hline
\hline
{}  & M=1\textsuperscript{$\dagger$} & M=2\textsuperscript{$\dagger$} & M=4\textsuperscript{$\dagger$} & M=6\textsuperscript{$\dagger$} & M=8\textsuperscript{$\dagger$} & M=10\textsuperscript{$\dagger$} & M=12\textsuperscript{$\dagger$} & M=14\textsuperscript{$\dagger$} & M=16\textsuperscript{$\dagger$} & M=18\textsuperscript{$\dagger$} \\ [0.5ex] % inserts table
\hline
resnet\_v1\_50  & 55.5 & 47.2 & 54.1 & 46.9 & 41.9 & 42.8 & 44.2 & 38.4 & 37.3 & 32.4 \\ % inserting body of the table
resnet\_v1\_101 & 49.7 & 54.6 & 40.2 & 51.2 & 44.9 & 57.2 & 44.7 & 41.7 & 42.2 & 34.5 \\
resnet\_v2\_50 & 42.3 & 49.7 & 52.9 & 40.8 & 42.3 & 43.8 & 57.8 & 40.4 & 40.9 & 35.2 \\
resnet\_v2\_101 & 54.4 & 38.9 & 49.2 & 44.9 & 41.5 & 44.8 & 44.02 & 42.0 & 39.6 & 50.6 \\
mobilenet\_v1\_100 & \textbf{60.5} & 55.8 & 51.2 & 49.8 & \textbf{ 47.7} & \textbf{45.3 } & 42.8 & \textbf{43.1} & 41.9 & \textbf{41.8}  \\
mobilenet\_v1\_075 & 58.2 & \textbf{ 57.9} & 52.4 & 51.1 & 46.9 & 44.1 & \textbf{ 45.2} & 41.8 & 41.2 & 40.2 \\
inception\_v1 & 51.3 & 54.4 & 45.8 & 44.9 & 42.5 & 41.2 & 44.8 & 38.8 & 35.3 & 45.8 \\
inception\_v2 & 44.2 & 38.2 & 42.4 & 49.4 & 45.9 & 44.3 & 41.0 & 42.5 & 39.4 & 47.1 \\ 
mobilenet\_v1\_050\textsuperscript{$\ddagger$} & 58.2 & 56.2 & \textbf{54.6} & 46.6 & 43.6 & 41.2 & 38.5 & 39.3 & 34.2 & 35.8   \\
mobilenet\_v1\_025\textsuperscript{$\ddagger$} & 54.8 & 54.3 & 52.9 & \textbf{52.5} & 43.5 & 44.7 & 41.1 & 42.6 & \textbf{42.5} & 38.5 \\ [1ex]
\hline %inserts single line
C3D & 81.7 & 82.6 & 78.0 & 82.8 & 82.2 & 82.1 & 83.5 & 83.1 & 82.6 & 82.3 \\ [1ex] % [1ex] adds vertical 
\hline %inserts single line
\end{tabular}
\newline \centering{\textbf{\textsuperscript{$\ddagger$}} stands for training from scratch instead of fine-tuning and \textbf{\textsuperscript{$\dagger$}} stands for budget model restarting.}
\newline 
\label{table:sbu-2fold-eval} % is used to refer this table in the text
\end{center}
\end{adjustwidth}
\end{table}

\section {Experiments on UCF-101 / VISPR}
\subsection{``Transferability'' Study of Privacy Attributes between UCF-101 and VISPR
}
\subsubsection{Selection of 17 and 7 Privacy Attributes}

\begin{figure}[htbp]
	\makebox [\textwidth][c]{
		\includegraphics[width=1.3\columnwidth]{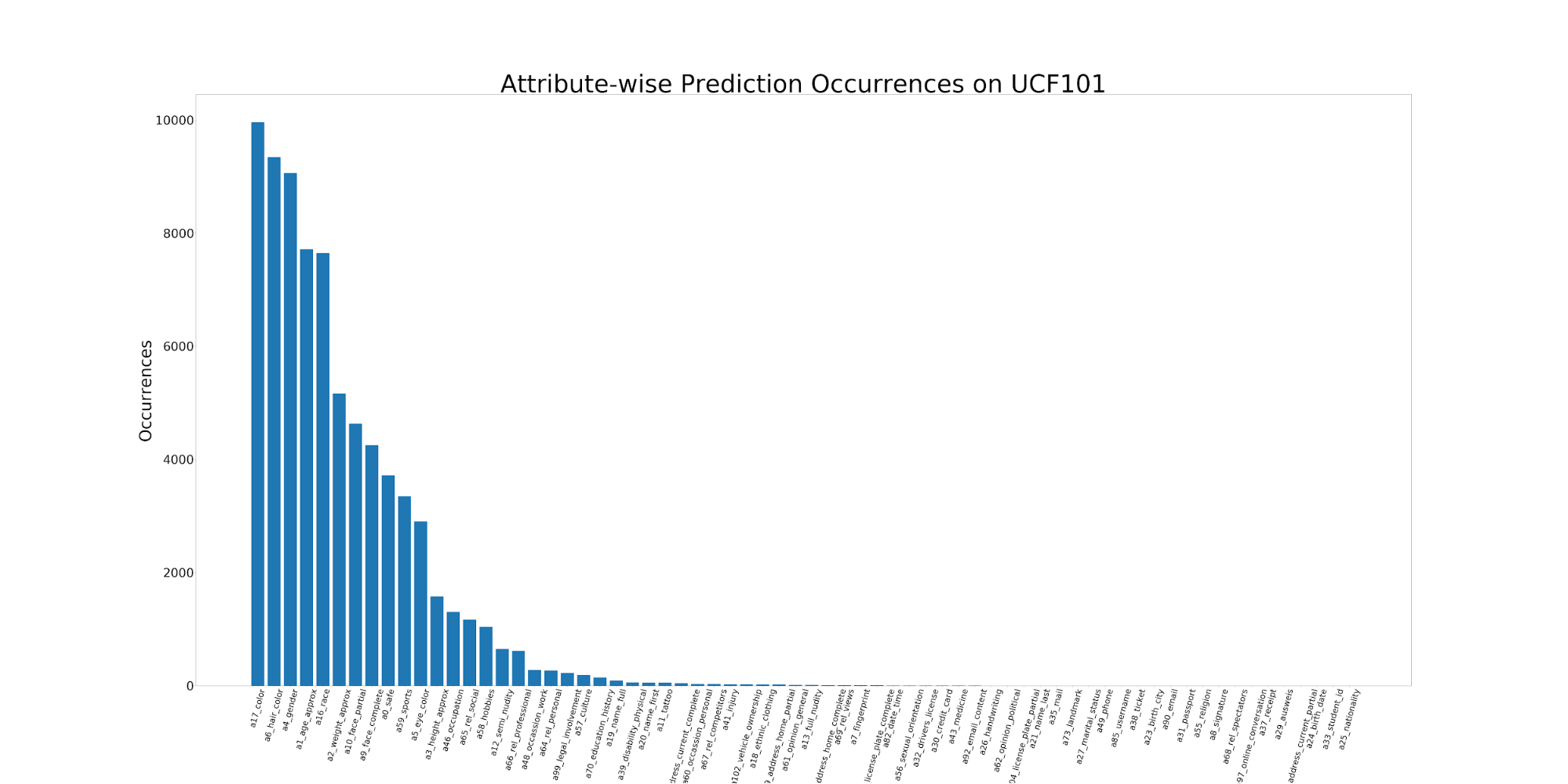}}
		
	\caption{Attribute-wise occurrence statistics on UCF-101 videos, evaluated using the pretrained privacy prediction model on VISPR.} 
	\label{fig:attribute}
\end{figure}

There are 13,421 videos in the UCF-101 dataset. For each video, we evaluate it using the privacy attribute prediction model pretrained on VISPR dataset: see the statistic plot in Figure \figref{attribute}, we observe that there are 43 attributes that can be found at least once in UCF101 videos. But only 17 out of the 43 are frequently occurring. These 17 attributes are \textit{\{age\_approx, weight\_approx, height\_approx, gender, eye\_color, hair\_color, face\_complete, face\_partial, semi-nudity, race, color, occupation, hobbies, sports, personal relationship, social relationship, safe\}}. 

Among the 17 frequent attributes, we carefully select 7 privacy attributes that best fit the smart home setting. These 7 attributes are \textit{\{semi-nudity, occupation, hobbies, sports, personal relationship, social relationship\}}.

\subsubsection{Privacy Attribute Examples in UCF-101}
\begin{figure}[htbp]
\begin{tabular}{cc}
\subfloat[ApplyLipStick]{\includegraphics[width = 2.5in]{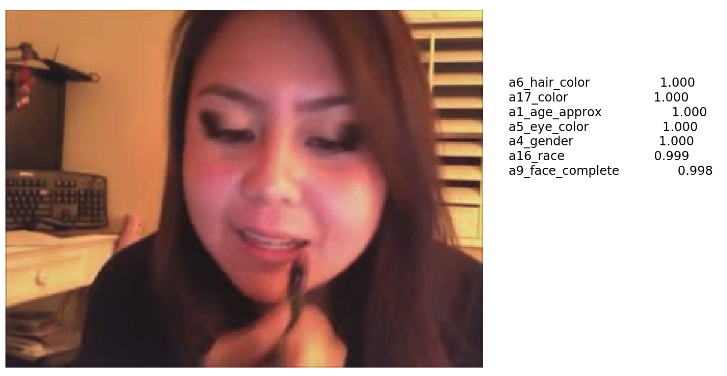}} &
\subfloat[BabyCrawling]{\includegraphics[width = 2.5in]{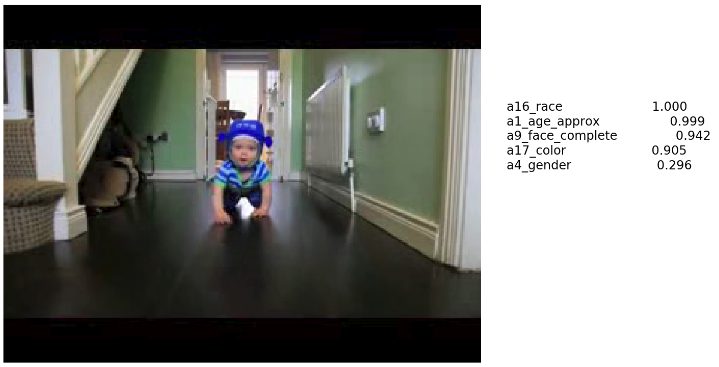}} \\
\subfloat[PlayingPiano]{\includegraphics[width = 2.5in]{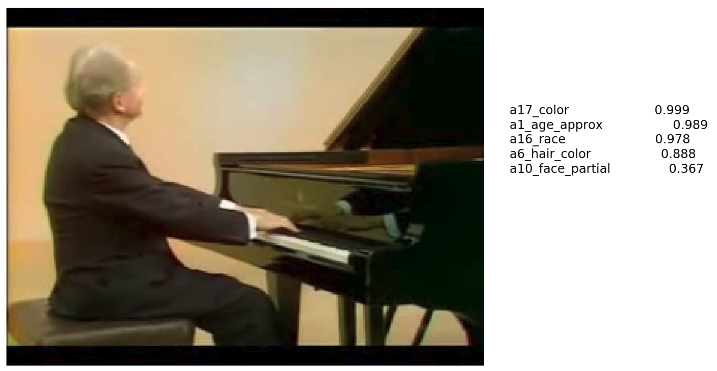}} & 
\subfloat[ShavingBeard]{\includegraphics[width = 2.5in]{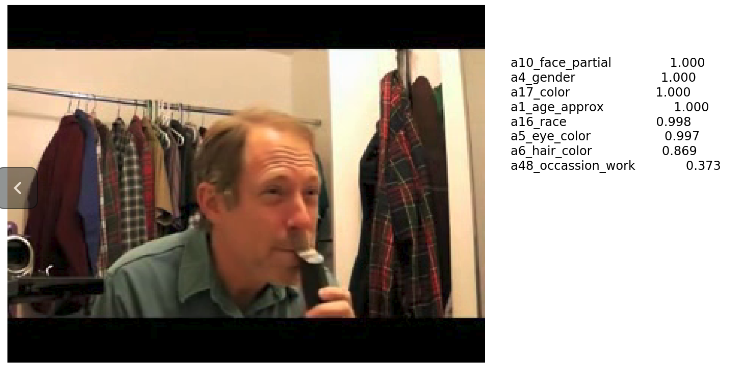}} \\
\subfloat[Situp]{\includegraphics[width = 2.5in]{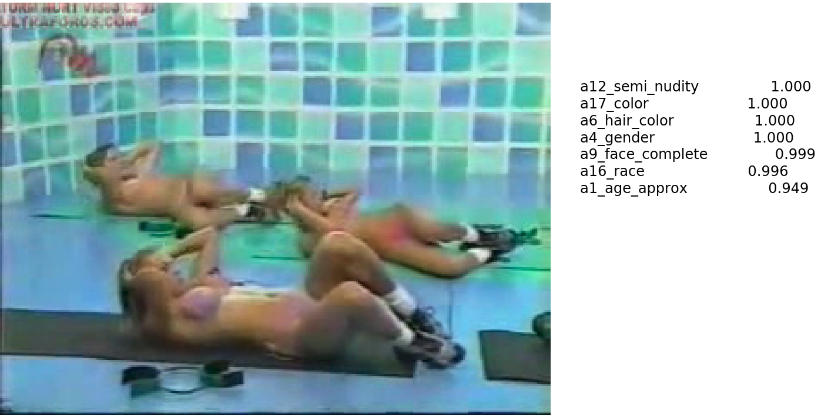}} &
\subfloat[YoYo]{\includegraphics[width = 2.5in]{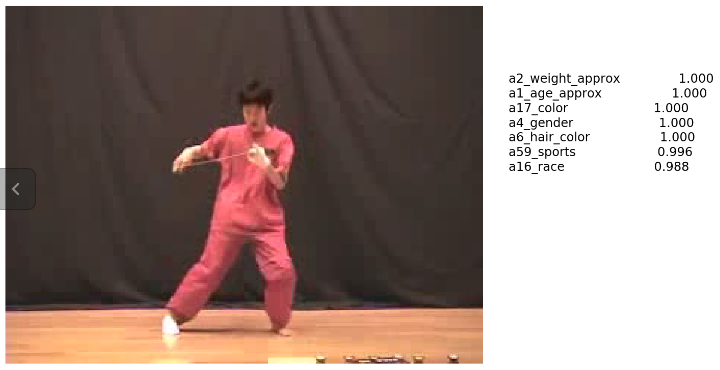}} \\
\end{tabular}
\caption{Privacy attributes prediction on example frames from UCF101. The right column denotes the predicted privacy attributes (as defined in the VISPR dataset \cite{orekondy2017towards}) and associated confidences from the left column frames, showing a high risk of privacy leak in daily common videos.}
	\label{fig:ucf_examples}
\end{figure}

In \figref{ucf_examples}, we show some example frames from UCF101 with privacy attributes predicted using the VISPR-pretrained model. In each example, the right column denotes the predicted privacy attributes (as defined in the VISPR dataset \cite{orekondy2017towards}) and associated confidences from the left column frames, showing a high risk of privacy leak in daily common videos. We qualitatively examine a large number of UCF-101 frames and determine that privacy attributes prediction are highly reliable.

\begin{figure}[htbp]
  \centering\includegraphics[width = 1.2in]{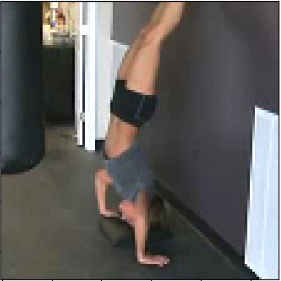}
  \caption*{Original RGB Frame from UCF-101 (Label: HandStandPushup)}
\captionsetup[subfigure]{labelformat=empty}
\begin{tabular}{cccc}
\subfloat[Method 2, M=1 \, \newline (VISPR-17)]{\includegraphics[width = 1.2in]{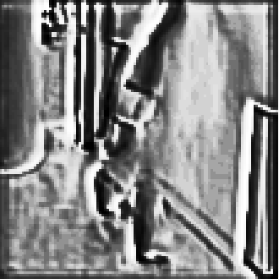}} &
\subfloat[Method 2, M=2 \, \newline  (VISPR-17)]{\includegraphics[width = 1.2in]{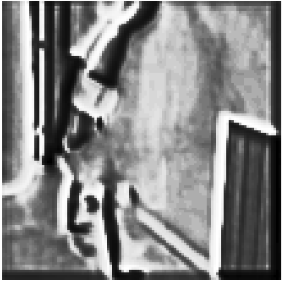}} &
\subfloat[Method 2, M=3 \, \newline  (VISPR-17)]{\includegraphics[width = 1.2in]{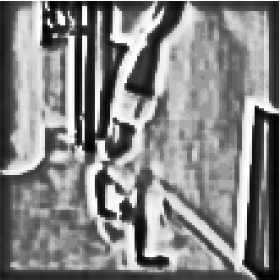}} &
\subfloat[Method 2, M=4 \, \newline  (VISPR-17)]{\includegraphics[width = 1.2in]{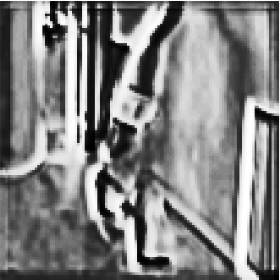}} \\
\subfloat[Method 3, M=1 \, \newline  (VISPR-17)]{\includegraphics[width = 1.2in]{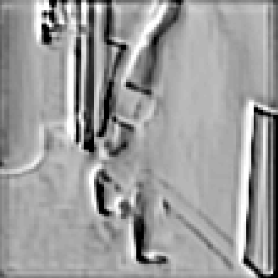}} &
\subfloat[Method 3, M=2 \, \newline  (VISPR-17)]{\includegraphics[width = 1.2in]{figures/ucf_visualization/17attributes_N_2_Restarting.png}} &
\subfloat[Method 3, M=3 \, \newline  (VISPR-17)]{\includegraphics[width = 1.2in]{figures/ucf_visualization/17attributes_N_3_Restarting.png}} &
\subfloat[Method 3, M=4 \, \newline  (VISPR-17)]{\includegraphics[width = 1.2in]{figures/ucf_visualization/17attributes_N_4_Restarting.png}} \\ 
\subfloat[Method 2, M=1 \, \newline  (VISPR-7)]{\includegraphics[width = 1.2in]{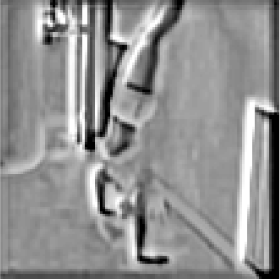}} &
\subfloat[Method 2, M=2 \, \newline  (VISPR-7)]{\includegraphics[width = 1.2in]{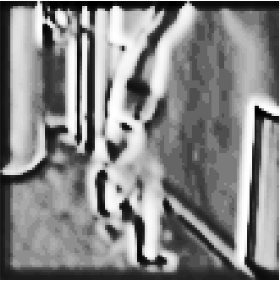}} &
\subfloat[Method 2, M=3 \, \newline  (VISPR-7)]{\includegraphics[width = 1.2in]{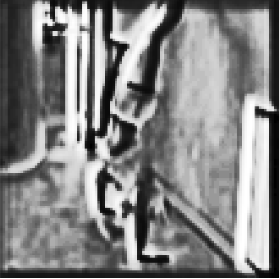}} &
\subfloat[Method 2, M=4 \, \newline  (VISPR-7)]{\includegraphics[width = 1.2in]{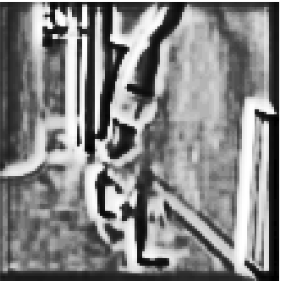}} \\
\subfloat[Method 3, M=1 \, \newline  (VISPR-7)]{\includegraphics[width = 1.2in]{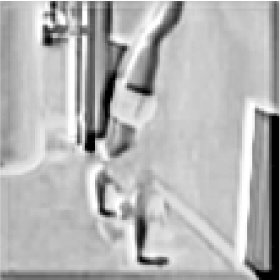}} &
\subfloat[Method 3, M=2 \, \newline  (VISPR-7)]{\includegraphics[width = 1.2in]{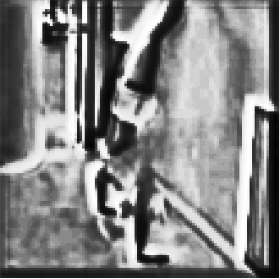}} &
\subfloat[Method 3, M=3 \, \newline  (VISPR-7)]{\includegraphics[width = 1.2in]{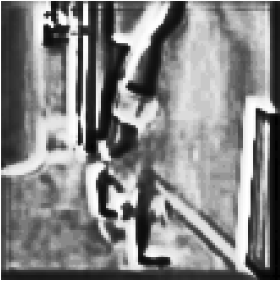}} &
\subfloat[Method 3, M=4 \, \newline  (VISPR-7)]{\includegraphics[width = 1.2in]{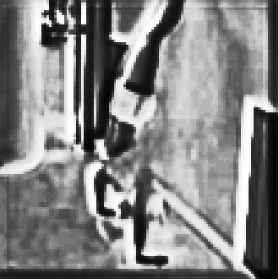}}
\end{tabular}
\caption{Example frames after applying the learned degradation on UCF-101 with adversarial training on VISPR-17 and VISPR-7}
	\label{fig:ucf_visualization}
\end{figure}

\subsection{UCF-101 / VISPR Two-Fold Evaluation}

The trade-off results between UCF-101 with VISPR-17 and VISPR-7 are found in Tables \ref{table:ucf-vispr-2fold-eval-17attributes} and \ref{table:ucf-vispr-2fold-eval-7attributes}, respectively. Note that for the $N$=10 privacy attribute prediction evaluation, the results are in class-based MAP (cMAP) rather than recognition accuracy. 

%We want to make an additional note that, in Table \ref{table:vispr-downsample} the privacy prediction is evaluated using \underline{only one model}; while in our adversarial training, the privacy suppression effect is evaluated using the \underline{highest achievable number} among 10 different models. Therefore, \textbf{the evaluation protocol for our adversarial training is ``stricter'', and its gain on privacy protection compared to downsampling will be essentially ``underestimated''}, if we just directly compare cMAPs in Table \ref{table:ucf-vispr-2fold-eval-7attributes} and \ref{table:ucf-vispr-2fold-eval-17attributes}, with Table \ref{table:vispr-downsample} in our paper: \textbf{the same for SBU}.

\begin{table}[htbp]
\caption{UCF-101 / VISPR-17 Two-Fold Evaluation} % title of Table
\centering % used for centering table
\begin{tabular}{|c|c|c|c|c|c|c|c|c|} % centered columns (4 columns)
\hline
{}  & M=1 & M=1\textsuperscript{$\dagger$} & M=2 & M=2\textsuperscript{$\dagger$} & M=3 & M=3\textsuperscript{$\dagger$} & M=4 & M=4\textsuperscript{$\dagger$}  \\ [0.5ex] % inserts table
\hline
resnet\_v1\_50  & \textbf{66.68} & 63.45 & 62.12 & 63.78 & \textbf{65.59} & 62.12 & \textbf{65.12} & 59.83 \\ % inserting body of the table
resnet\_v1\_101 & 65.78 & 59.24 & 62.48 & 61.29 & 59.59 & 61.23 & 64.21 & 61.49  \\
resnet\_v2\_50 & 62.12 & \textbf{65.28 } & \textbf{66.94} & 62.48 & 59.59 & 59.56 & 62.34 & 60.47 \\
resnet\_v2\_101 & 59.12 & 61.45 & 57.59 & 59.43 & 58.32 & 61.43 & 64.23 & 59.48  \\
mobilenet\_v1\_100 & 63.45 & 58.48 & 62.69 & 61.47 & 64.39 & 61.59 & 65.01 & 57.43 \\
mobilenet\_v1\_075 & 62.23 & 62.48 & 64.28 & 59.47 & 60.27 & 58.57 & 55.48 & 57.57  \\
inception\_v1 & 58.32 & 62.49 & 59.39 & \textbf{64.82} & 63.57 & 61.39 & 63.58 & 58.46  \\
inception\_v2 & 65.79 & 61.28 & 64.52 & 63.58 & 60.49 & \textbf{63.58} & 60.25 & 59.39  \\ 
mobilenet\_v1\_050\textsuperscript{$\ddagger$} & 65.12 & 60.25 & 64.29 & 59.49 & 62.48 & 63.58 & 63.58 & \textbf{62.06}  \\
mobilenet\_v1\_025\textsuperscript{$\ddagger$} & 62.54 & 63.59 & 62.58 & 62.46 & 60.47 & 59.20 & 58.27 & 61.36 \\ [1ex]
\hline %inserts single line
C3D & 66.58 & 66.36 & 64.46 & 65.27 & 65.28 & 65.89 & 66.59 & 65.83     \\ [1ex] % [1ex] adds vertical 
\hline %inserts single line
\end{tabular}
\newline \textbf{\textsuperscript{$\ddagger$}} stands for training from scratch instead of fine-tuning and \textbf{\textsuperscript{$\dagger$}} stands for budget model restarting
\label{table:ucf-vispr-2fold-eval-17attributes} % is used to refer this table in the text
\end{table}
\begin{table}[!h]
\caption{UCF-101 / VISPR-7 Two-Fold Evaluation} % title of Table
\centering % used for centering table
\begin{tabular}{|c|c|c|c|c|c|c|c|c|} % centered columns (4 columns)
\hline
{}  & M=1 & M=1\textsuperscript{$\dagger$} & M=2 & M=2\textsuperscript{$\dagger$} & M=3 & M=3\textsuperscript{$\dagger$} & M=4 & M=4\textsuperscript{$\dagger$}  \\ [0.5ex] % inserts table
\hline
resnet\_v1\_50  & \textbf{40.68} & 38.24  & 38.45 & 35.67 & 35.34 &32.54 & 35.58 & 33.41  \\ % inserting body of the table
resnet\_v1\_101 & 32.21 & 37.69 & 37.31 & 36.21 & 37.35 & 34.53 & 37.48 & 32.67 \\
resnet\_v2\_50 & 33.46 & 37.13 & \textbf{39.94} & \textbf{36.28} &32.59 & 34.13 & 36.69 & 33.46 \\
resnet\_v2\_101 & 35.25 & 34.49 & 32.58 & 35.38 & \textbf{38.59} & \textbf{35.16} & 37.24 & 31.53  \\
mobilenet\_v1\_100 & 33.28 & 35.24 & 37.54 & 32.48 & 31.59 & 28.36 & 32.48 & 29.57 \\
mobilenet\_v1\_075 & 28.59 & 34.58 & 38.23 & 31.59 & 35.38 &  30.94 & 29.58 & 32.58 \\
inception\_v1 & 35.28 & 37.56 & 36.84 & 27.48 & 29.48 & 30.48 & 32.04 & \textbf{34.48}  \\
inception\_v2 & 38.47 & 36.39 & 35.29 & 30.92 & 28.59 & 33.59 & 35.38 & 29.58  \\ 
mobilenet\_v1\_050\textsuperscript{$\ddagger$} & 38.49 & 28.49 & 32.56 & 33.48 & 31.58 & 32.58 & \textbf{38.32} & 33.48 \\
mobilenet\_v1\_025\textsuperscript{$\ddagger$} & 35.47 & \textbf{38.42} & 34.93 & 31.28 & 33.37 & 34.78 & 33.57 & 30.08 \\ [1ex]
\hline %inserts single line
C3D   & 65.16  & 65.58 & 64.53 & 66.46  & 65.38 & 64.28 & 64.83 & 65.37    \\ [1ex] % [1ex] adds vertical 
\hline %inserts single line
\end{tabular}
\newline \textbf{\textsuperscript{$\ddagger$}} stands for training from scratch instead of fine-tuning and \textbf{\textsuperscript{$\dagger$}} stands for budget model restarting
\label{table:ucf-vispr-2fold-eval-7attributes} % is used to refer this table in the text
\end{table}

\subsection{Visualization Examples of Learned Degradation on UCF-101 / VISPR}
For visualized examples of learned $f_d(X)$, please refer to \figref{ucf_visualization} for VISPR-17 and VISPR-7.

\end{subappendices}

\end{document}